# DedustNet: A Frequency-dominated Swin Transformer-based Wavelet Network for Agricultural Dust Removal

Shengli Zhang[a,1], Zhiyong Tao[a,**,2] and Sen Lin[b,3]

[a]*School of Electronic and Information Engineering, Liaoning Technical University, Huludao, 125105, Liaoning, China*
[b]*School of Automation and Electrical Engineering, Shenyang Ligong University, Shenyang, 110159, Liaoning, China*

ABSTRACT

While dust significantly affects the environmental perception of automated agricultural machines, it still needs to be explored, and the existing deep learning-based methods for dust removal require further refinement and development. Further research and improvement in this area are essential to improve the performance and reliability of automated agricultural machines in dusty environments. As a solution, we propose an end-to-end trainable learning network (DedustNet) to solve the real-world agricultural dust removal task. To our knowledge, DedustNet is the first time Swin Transformer-based units have been used in wavelet networks for agricultural image dusting. Specifically, we present the frequency-dominated Swin Transformer-based block (DWTFormer block and IDWTFormer block) by adding a spatial features aggregation scheme (SFAS) to the Swin Transformer and combining it with the wavelet transform. As the basic blocks of the encoding and decoding, the DWTFormer block and IDWTFormer block recover richer details, such as the structure and texture of the image, alleviating the limitation of the global receptive field of Swin Transformer when dealing with complex dusty backgrounds. Furthermore, We propose a cross-level information fusion module (CIFM) to fuse different levels of features and effectively capture global and long-range feature relationships. In addition, we present a dilated convolution module (DCM) to capture contextual information guided by wavelet transform at multiple scales, which combines the advantages of wavelet transform and dilated convolution. Our algorithm leverages deep learning techniques to effectively remove dust from images captured in real-world agricultural settings while preserving the original structural and textural features. Compared to existing state-of-the-art methods, DedustNet achieves superior performance and more reliable results in agricultural image dedusting, providing strong support for the application of agricultural machinery in dusty environments. Additionally, the impressive performance on real-world hazy datasets and application tests highlights DedustNet's superior generalization ability and computer vision-related application performance compared to state-of-the-art methods.

## 1. Introduction

Dust is a common phenomenon that can significantly reduce the quality of captured images. This has significant implications for the performance of automated equipment, especially in the context of mechanized agricultural work. Therefore, single image dedusting is a crucial low-level vision task, posing a significant challenge in agricultural landscape restoration tasks. It is essential to address this issue to improve the quality and applicability of automated equipment in agriculture.

Dedusting techniques typically utilize methods in image processing (Jiang, Wang, Yi, Jiang, Xiao and Yao (2018); Jiang, Wang, Yi, Wang, Lu and Jiang (2019)) and pattern recognition (Rasti, Uiboupin, Escalera and Anbarjafari (2016); Wang, Yi, Jiang, Jiang, Han, Lu and Ma (2018)) to identify and weaken noise and pollutants in images, thereby removing them and restoring the original features of the picture. This can also improve images' visual clarity and color balance, enhancing their aesthetic appeal and readability. As such, dedusting techniques have wide-ranging applications in the agricultural field.

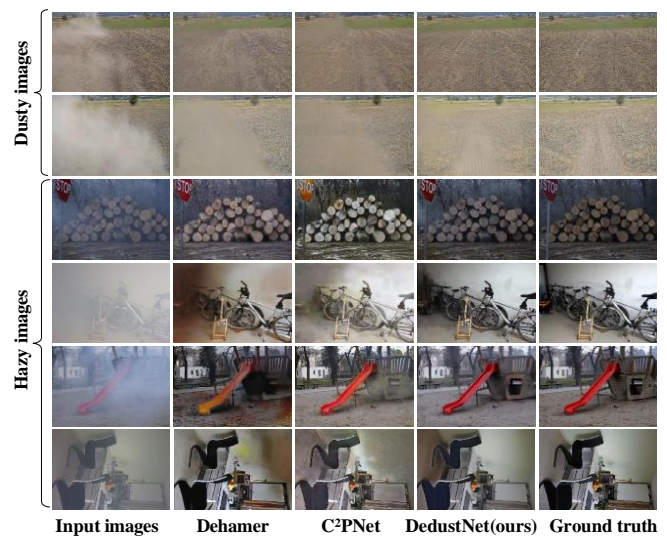

**Figure 1:** Qualitative comparison among DedustNet and SOTA methods on image dedusting and dehazing on RB-Dust dataset and hazy benchmark datasets.

The RB-Dust dataset(Buckel, Oksanen and Dietmueller (2023)) has been proposed recently for agricultural landscape dust removal. This work also validated the effectiveness of image dehazing methods for dust removal. The classical atmospheric scattering model (ASM) (Wang, Huang,

**Zhiyong Tao. School of Electronic and Information Engineering, Liaoning Technical University, Huludao 125105, China. (Z.

✉ zhangshengli_win@163.com (S. Zhang); taozhiyong@lntu.edu.cn
Tao); lin_sen6@126.com (S. Lin)
ORCID(s):

Zou and Xu (2021a); McCartney (1976); Narasimhan and Nayar (2002); Nayar and Narasimhan (1999)), which has been employed by formal dehazing methods to characterize the process of fog image degradation by:

$$I(x) = J(x)t(x) + A(1 - t(x)) \quad (1)$$

where $I(x)$ and $J(x)$ are the degraded images and clear images respectively. $A$ represents global atmospheric light, $t(x) = e^{-\beta d(x)}$ is transmission map, where $\beta$ and $d(x)$ represent the atmospheric scattering parameters and scene depth, respectively.

However, these prior-based methods (He, Sun and Tang (2010); Zhu, Mai and Shao (2015); Li, Peng, Wang, Xu and Feng (2017); Fattal (2014); Zhang, Wang, Yang, Zhang, He and Song (2018); Zhou, Teng, Han, Xu and Shi (2021)) often involve time-consuming iterative optimization and manually designed priors, which may not always align with practical scenarios. Dust formation is influenced by various natural factors, such as temperature and humidity, making it challenging to model simplistically and accurately. As a result, these prior-based methods may lead to estimation errors when dealing with complex scenes.

In recent years, data-driven methods (Ren, Liu, Zhang, Pan, Cao and Yang (2016b); Dong, Pan, Xiang, Hu, Zhang, Wang and Yang (2020a); Singh, Bhave and Prasad (2020); Yang, Yang and Tsai (2020); Das and Dutta (2020)) have shown significant progress in image processing. Several end-to-end models (Cai, Xu, Jia, Qing and Tao (2016); Liu, Ma, Shi and Chen (2019); Ren, Ma, Zhang, Pan, Cao, Liu and Yang (2018); Wang, Yu, An and Wei (2021b); Yang et al. (2020); Das and Dutta (2020)) have been proposed to reduce the reliance on predefined prior information, but they may have limited interpretability. With the advancements in deep learning techniques and the availability of large-scale synthetic datasets (Liu, Zhu, Pei, Fu, Qin, Zhang, Wan and Feng (2021a)), many data-driven approaches (Wu, Qu, Lin, Zhou, Qiao, Zhang, Xie and Ma (2021); Fu, Liu, Yu, Chen and Wang (2021); Guo, Yan, Anwar, Cong, Ren and Li (2022); Li, Li, Zhao and Xu (2021); Zheng, Zhan, He, Dong and Du (2023); Song, He, Qian and Du (2023)) utilize Convolutional Neural Networks (CNNs) to extract features and build end-to-end dehazing networks that learn the transmission maps between clear and degraded images.

However, current image enhancement methods still have the following problems: (a) Swin Transformer (Liu, Lin, Cao, Hu, Wei, Zhang, Lin and Guo (2021b)) uses shifted windows to contribute hierarchical feature maps but suffers from a limitation of global information and receptive field due to the uneven distribution of dust density. (b) The distribution of dust in the real world is complex, and downsampling can lead to color distortion or loss of detail in output results as the resolution of a feature map or image resolution decreases; (c) Improvements are still needed to balance the generalization ability and robustness of networks with the complexity of models when addressing the challenging visual tasks.

We all know that the low-frequency layer of an image captures more structural information, such as color and target. In contrast, the high-frequency layer represents specific details (e.g., edges, textures). This frequency information is crucial for reconstructing the structure and texture of an image. Therefore, we leverage the discrete wavelet transform (DWT) and inverse discrete wavelet transform (IDWT) to decompose the RGB image into high and low-frequency information, which can guide the network in image recovery. This approach also helps to avoid information loss and increase the receptive field, achieving a better balance between efficiency and restoration performance.

Combined analysis of the above, we present DedustNet and summarize the contributions as follows:

1) As shown in 3, we propose a frequency-dominated (DedustNet) for agricultural dust removal. To our knowledge, this is the first time Swin Transformer-based units are used in wavelet networks for agricultural image dusting in the real world.

2) We design DWTFormer and IDWTFormer Block, combining the advantages of wavelet transform and Swin Transformer with spatial features aggregation scheme (SFAS) to recover details such as structure and texture of the image guided by frequency information, and improve the overall receptive field of the network under complex dusty backgrounds.

3) We propose the cross-level information fusion module (CIFM) to fuse different levels of features, recovering rich texture details and information and effectively capturing global self-similarity and long-range feature relationships. The dilated convolution module (DCM) is proposed to capture contextual information guided by wavelet transform at multiple scales, combining the advantages of wavelet transform and dilated convolution.

4) DedustNet has shown superior performance and more reliable results in image dedusting compared to existing state-of-the-art methods, further supporting the application of agricultural machinery in dusty environments.

5) DedustNet outperforms state-of-the-art methods in qualitative and quantitative; the satisfactory results obtained on the dense and non-homogeneous dehazing task also demonstrate the robustness and generalization capability of the DedustNet. DedustNet's performance in the application test also exhibits its superior performance in computer vision-related applications.

The remaining sections of this paper are structured as follows. Section 2 provides a concise overview of the existing research on image dehazing, wavelet transform, and ViT. In Section 3, we outline the proposed DedustNet framework, its implementation details, and the loss function we adopted. We present the dataset details and experimental setup we used in Section 4. The evaluation of our proposed method is shown in Section 5, including a comparison to SOTA methods, generality analysis for DedustNet, application test, and ablation study. Finally, Section 6 and Section 7 summarize the conclusions and discussions derived from our study.

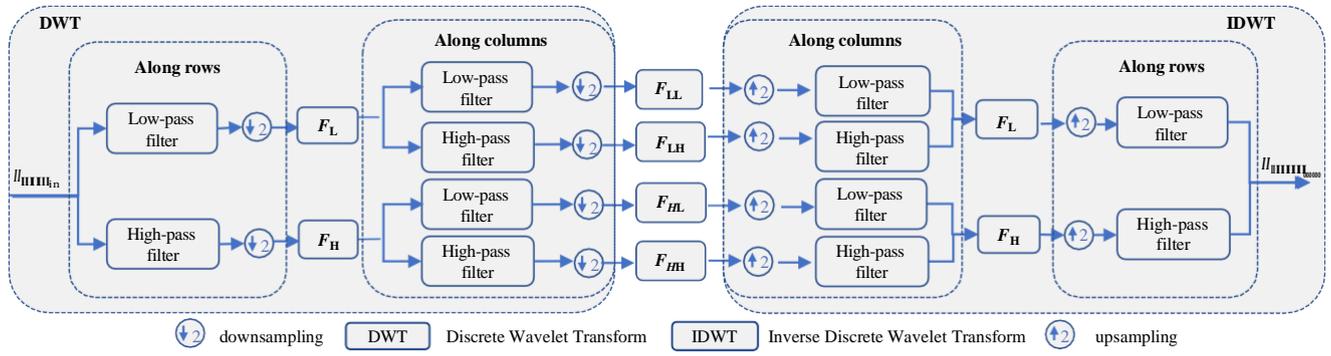

**Figure 2:** Example of the 2D discrete Wavelet transform.

## 2. Related work

Image dedusting is currently a significant task that has yet to receive much attention, and in this work (Buckel et al. (2023)), algorithms similar to image dehazig and related image enhancement algorithms can perform the task of image dedusting. Therefore, we next focus on related work in image dehazing in Senction 2.1. We also introduce the related work of discrete wavelet transform in Section 2.2 and Vision Transformer in Section 2.3.

### 2.1. Image dehazing

The existing methods for image dehazing are broadly classified into two categories: traditional prior-based methods and learning-based methods.

#### 2.1.1. Traditional methods

Most prior-based dehazing methods (He et al. (2010); Zhu et al. (2015); Li et al. (2017); Fattal (2014); Zhang et al. (2018); Zhou et al. (2021)) use hazy and clear images to estimate the transmission map, then use ASM to recover haze-free images. The dark channel prior (DCP) (He et al. (2010)) is proposed to assume that the image patches of haze-free outdoor images often have low-intensity values in at least one channel. To address the difference in brightness and saturation of hazy images, the color attenuation prior (CAP) (Zhu et al. (2015)) is proposed to estimate the scene depth as solid prior knowledge. To eliminate the polarization effect of information, a globally nonuniform ambient light model (Shen, Zhang, Li, Yuan and Zhang (2020)) is proposed to predict spatially varied ambient light and designed a bright pixel index to correct the transmission. With the predicted haze parameters, they reversed the atmospheric scattering model to restore visibility. Therefore, a robust polarization-based dehazing network (Zhou et al. (2021)) is proposed. However, the performance of these methods is inherently limited by the specific scenario, and they may lead to undesirable color distortions when the scenario does not satisfy these priors. In contrast, DedustNet can reconstruct images with richer detail by leveraging the complementary advantages of prior- and deep learning-based methods.

#### 2.1.2. Deep learning methods

Recently, deep learning techniques have been proposed to tackle the problem of underwater image dehazing. These techniques have shown promising results in the restoration of underwater images. They can be classified into two categories: (i) CNN-based methods, (ii) GAN-based methods.

**CNN-based methods.** CNN-based methods Ren et al. (2016b); Cai et al. (2016); Li et al. (2017); Liu et al. (2019); Qin, Wang, Bai, Xie and Jia (2020); Singh et al. (2020); Wu et al. (2021); Song et al. (2023); Zheng et al. (2023) have dominated in recent years. The MSCNN (Ren et al. (2016b)) is proposed to estimate $t(x)$ using a coarse-scale network followed by local optimization. AODNet (Li et al. (2017)) is presented to learn each hazy image and its $t(x)$, which reiterated ASM. However, all of these methods rely on ASM, and the dehazing results are often color-distorted. To alleviate the bottleneck problem encountered in traditional multi-scale methods, the GridDehazeNet (Liu et al. (2019)) is proposed, which implemented an attention-based end-to-end dehazing network. To enable more efficient dehazing network performance, the FFANet (Qin et al. (2020)) is presented with channel and spatial attention to obtain excellent dehazing performance. By taking FFANet as a baseline, the C$^2$PNet (Zheng et al. (2023)) is proposed with a curricular contrastive regularization and the physics-aware dual-branch unit to enhance the network dehazing performance. However, behind the excellent performance achieved by these supervised methods, a large number of data pairs are required for the training; more importantly, these methods are almost trained on synthetic images, which can not be well generalized to real-world image dehazing.

**GAN-based methods.** Recently, some unsupervised data-driven methods (Ren, Liu, Zhang, Pan, Cao and Yang (2016a); Mehta, Sinha, Mandal and Narang (2021); Dong, Liu, Zhang, Chen and Qiao (2020b); Fu et al. (2021); Wang, Zhu, Huang, Zhang and Wang (2022); Li, Zheng, Shu and Wu (2022)) have also made significant progress in image defogging. The Generative Adversarial Networks (GANs) is first introduced to the field of image defogging and proposed an end-to-end defogging network that achieves mapping from foggy images to fog-free images by training a generator and a discriminator. The SkyGAN (Mehta et al. (2021)) is presented for haze removal in aerial images, alleviating the degradation in image visibility. Recently, an end-to-end GANs with Fusion discriminator (FD-GAN) (Dong et al. (2020b)) is proposed for image dehazing, this model can generator more natural and realistic dehazed images with less color distortion and fewer artifacts. To address the issuse between trained synthetic blurred images and untrained real blurred images, a dual multiscale network, TMS-GAN (Wang et al. (2022)) is proposed to alleviate the problem of limited domain transfer performance. To reduce the number of training data significantly, a novel single image dehazing algorithm (Li et al. (2022)) is presented by combining model-based and data-driven approaches, and the proposed neural augmentation framework converges faster than the corresponding data-driven approach. However, unsupervised methods may be unstable during the training process, leading to problems such as the possibility of unstable results during enhancements. .

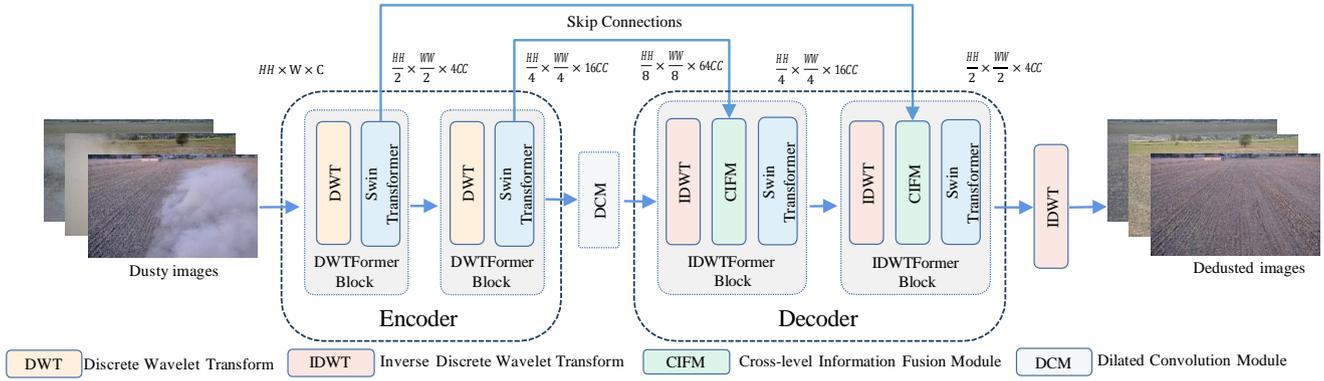

**Figure 3:** Schematic illustration of our DedustNet. The difference between encoding and decoding is that downsampling and upsampling are replaced by discrete wavelet transform (DWT) and inverse discrete wavelet transform (IDWT).

### 2.2. Discrete wavelet transform

Fig. 2 illustrates the overall schematic of the DWT and IDWT. As a traditional image processing technique, the discrete wavelet transform Claypoole, Baraniuk and Nowak (1998); Yang et al. (2020); Das and Dutta (2020); Liu, Yan and Zhao (2020) is widely used for image analysis. The DWSR (Guo, Seyed Mousavi, Huu Vu and Monga (2017)) is presented to combine the discrete wavelet transform with ResNet by predicting the residual wavelet subbands. Inspired by U-Net, the MWCNN (Liu, Zhang, Zhang, Lin and Zuo (2018)) is proposed, which replaces pooling and non-pooling operation to reduce the number of parameters in the network. However, multiple uses of the discrete wavelet transform operations may result in redundant channels. Therefore, the Wavelet U-Net (Yang and Fu (2019)) is proposed to use the discrete wavelet transform to extract edge features while applying the adaptive color transform that convolutional layers; this structure enhances the texture details in the image. To obtain large sensory fields with a high spatial resolution, the SDWNet (Zou, Jiang, Zhang, Chen, Lu and Wu (2021)) is proposed and recovers precise high-frequency texture details. Furthermore, a two-branch network DW-GAN (Fu et al. (2021)) is presented to leverage the power of discrete wavelet transform in helping the network acquire more frequency domain information. These methods demonstrate the significant role of discrete wavelet transform in the image recovery process.

### 2.3. Vision Transformer

Attention mechanism (Vaswani, Shazeer, Parmar, Uszkoreit, Jones, Gomez, Kaiser and Polosukhin (2017)) of deep learning has achieved a great process nowadays. Recently, Transformer (Vaswani et al. (2017)) has gained increasing attention, image content and attention weights interact spatially as a result of spatially varying convolution. The Vision Transformer (ViT) (Dosovitskiy, Beyer, Kolesnikov, Weissenborn, Zhai, Unterthiner, Dehghani, Minderer, Heigold, Gelly et al. (2020)) is proposed with the direct application of the Transformer architecture, which projects images into token sequences via patch-wise linear embedding. The shortcomings of the ViT are its weak inductive bias and its quadratic computational cost. Until the Swin Transformer (Liu et al. (2021b)) is presented, they divided tokens into a window and performed self-attention within a window to maintain a linear computational cost. Dehamer (Guo et al. (2022)) is proposed to modulate convolutional features via learning modulation matrices, which are conditioned on Transformer features instead of simple addition or concatenation of features. Dehamer effectively integrates Transformer features and CNN features and bring the domain knowledge such as task-specific prior into Transformer for improving the performance. Furthermore, the Dehaze-Former (Song et al. (2023)) is presented to improve on Swin Transformer Liu et al. (2021b), which can be viewed as a combination of Swin Transformer and U-Net Ronneberger, Fischer and Brox (2015) with more comprehensive improvements in the normalization layer, nonlinear activation function, and spatial information aggregation scheme. DehazeFormer improves the network performance for single-image dehazing further. Although ViT enhances the image recovery performance, it may increase additional computational expense and ignore the haze distribution characteristics under complex conditions.

## 3. Methodology

In the section, We show in Section 3.1 the motivation for the proposed DedustNet and the main components of our network. We show our proposed DWTFormer and IDWT-Former blocks in Section 3.2. In Section 3.3 and Section 3.4, we introduce the proposed CIFM and DCM, respectively. Finally, we present the loss function adopted in this paper in Section 3.5.

### 3.1. Motivation for DedustNet

Fig. 3 illustrates the framework of our proposed Dedust-Net. The encoding and decoding of DedustNet are based on the DWTFormer and IDWTFormer blocks. Although the DWT-Former block as the base block of the network mainly combines wavelet transform and Swin Transformer, we do not directly apply these existing tools but improve them. We use the wavelet transform to transform the features to the frequency domain and use the frequency information to guide DedustNet to recover the structural and texture details of the image. In addition, our proposed spatial features aggregation scheme (SFAS) with parallel convolution also alleviates the receptive field caused by Swin Transformer. This structure of the proposed DWTFormer block also alleviates the details caused by downsampling loss and other problems. Furthermore, we combine a cross-level information fusion module (CIFM) to integrate information from two different encoding and decoding stages, which can effectively capture global self-similarity and long-range feature relationships. To connect encoding and decoding, we propose the dilated convolution module (DCM) that serves as an interface between the two stages to complete the feature interaction in different receptive fields.

### 3.2. DWTFormer and IDWTFormer blocks

DWTFormer Block primarily uses DWT to convert information to wavelet domain from the spatial domain, then puts it into the Swin Transformer with SFAS for global processing. The difference between the encoding and decoding is that downsampling and upsampling are replaced by DWT and IDWT, respectively.

#### 3.2.1. Frequency subband decomposition

For the 2D discrete wavelet transform, we import the PyWavelets library and use Daubechies wavelet basis functions to satisfy tensor calculus and automatic gradient descent. Fig. 4 illustrates the network structure diagram of the DWTFormer block. We can observe that the dusty image is first decomposed into four frequency bands by DWT, then the output feature is input into Swin Transformer for feature processing at the same time, the convolution-operated is concatenated with the output of Multi-Head Self-Attention (MHSA) (Vaswani et al. (2017)). The output of DWTFormer Block is the sum of the outputs.

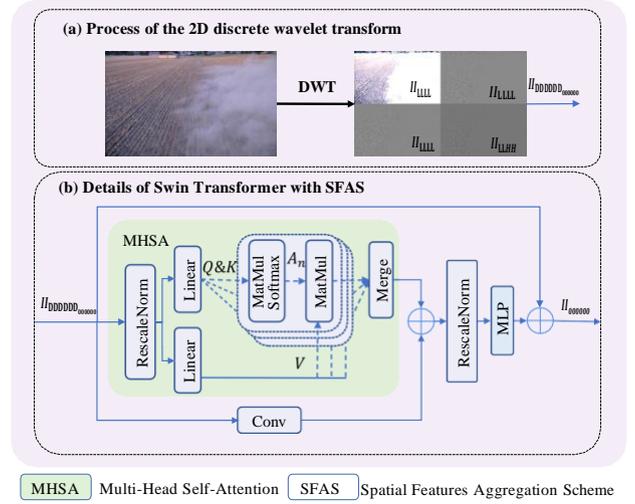

**Figure 4:** Architecture of proposed DWTFormer Block. The difference between the encoding and decoding is that down-sampling and upsampling are replaced by DWT and IDWT, respectively.

#### 3.2.2. Spatial features aggregation scheme

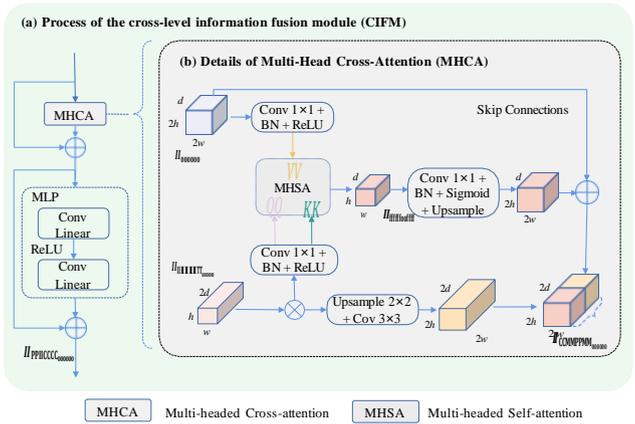

**Figure 5:** Architecture of proposed cross-level information fusion module (CIFM).

According to the attention mechanism (Vaswani et al. (2017)), given an input feature map (query, key, value), we compute the attention function for a set of queries simultaneously and pack them into a matrix. Swin Transformer applies MHSA within the window, and the MHSA and the self-attention of Swin Transformer can be expressed as follows:

Swin Transformer uses shifted windows to contribute hierarchical feature maps and solve the problem of not being able to transfer information from window to window, but Swin Transformer suffers from a limitation of global information and receptive field due to the uneven distribution of dust density. Therefore, we introduce SFAS into the DWTFormer Block in Fig. 4. In particular, we perform a further convolution of features from the DWT, DWTFormer Block realizes a dynamic style of information aggregation

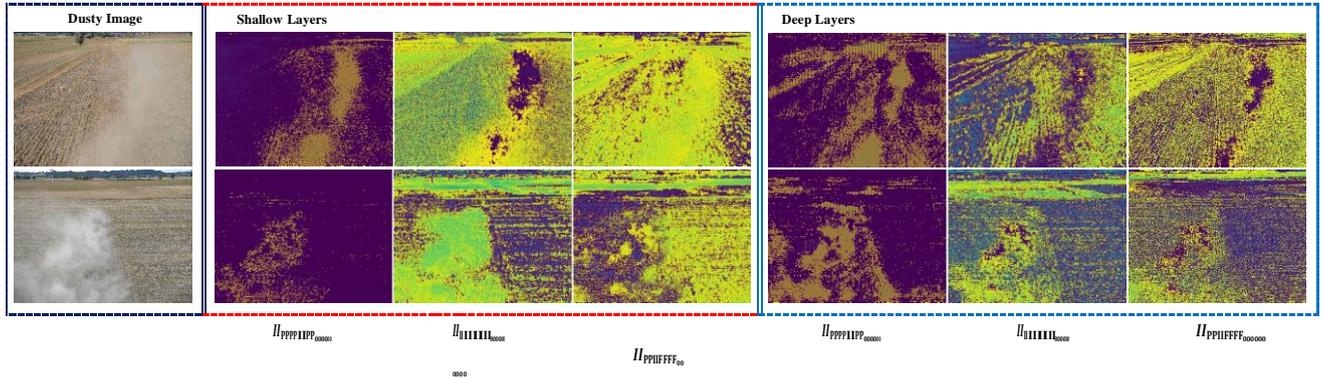

**Figure 6:** Visual results of the intermediate features in our proposed cross-level information fusion module. The corresponding modulated features are also presented. The features of the DWTFormer block in the proposed DWTFormer block have long-range attention but coarse textures, while the decoding features have clear details. The modulated features produced by the feature aggregation module inherit the characteristics of both Transformer features and frequency information, i.e., long-range dependencies and clear textures.

with MHSA generation in the spatial dimension. The output from the DWTFormer Block can thus be expressed as:

### 3.3. Cross-level information fusion module

Fig. 5 illustrates the framework of our proposed CIFM. The main network of DedustNet is a coding and decoding network composed of wavelet transform and Swin-Transformer. CIFM links the encoding and decoding stages, guiding DedustNet to generate images with more textures and rich details. We use Multi-Head Cross-Attention (MHCA) (Petit, Thome, Rambour, Themyr, Collins and Soler (2021)) to interoperate the information of the encoding and decoding stages. MHCA aims to highlight those critical areas for the application by removing unneeded or distracting areas from the skip connection features. MHCA performs a high-level mapping of the two inputs of the DWTFormer block. The obtained weight of MHCA is rescaled between 0 and 1 by the sigmoid function to obtain output. We regard noisy or irrelevant features as low-amplitude elements and filter out these disturbances. Finally, we connect the output with the high-level feature vector to obtain the output.

Fig. 6 illustrates the visual results of the intermediate features in our proposed CIFM. The input and output deep feature maps of CIFM possess more precise structural and textural information than shallow feature maps. This also shows that our proposed CIFM can fully retain the details of high-dimensional features and capture the long-range relationship among different receptive field features, thus improving decoding efficiency and enhancing the expressive ability of our DedustNet.

### 3.4. Dilated convolution module

The atrous spatial pyramid pooling module (ASPP) (Chen, Zhu, Papandreou, Schroff and Adam (2018)) is used in semantic segmentation tasks to capture contextual information at multiple scales. It involves using atrous convolutions with varying dilation rates to extract features at different receptive fields. This allows the model to understand an image's context and semantics better, leading to improved segmentation performance.

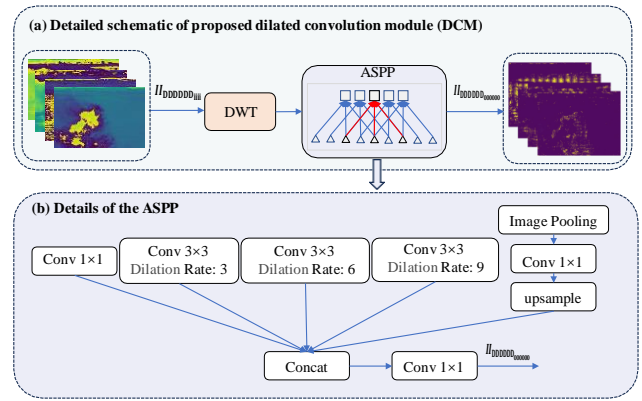

**Figure 7:** Architecture of our adopted dilated convolution module (DCM).

Fig. 7 illustrates the principle of the proposed DCM in our paper, which combines DWT and atrous spatial pyramid pooling module (ASPP module) (Chen et al. (2018)). The feature maps are input into the ASPP module after the frequency decomposition of the feature map via DWT. Unlike previous dehazing networks that use repeated upsampling and downsampling to obtain large receptive domains, we use dilated convolution with different expansion rates.

### 3.5. Loss founction

Referring to previous work (Zhao, Gallo, Frosio and Kautz (2016)), to balance both visual perception and quantitative assessments, we combine $\ell_1$ loss, multiscale structural similarity (MS-SSIM) loss, and perceptual loss linearly. Concretely, the $\ell_1$ loss retains color and brightness and converges quickly, providing a broader and more stable gradient.

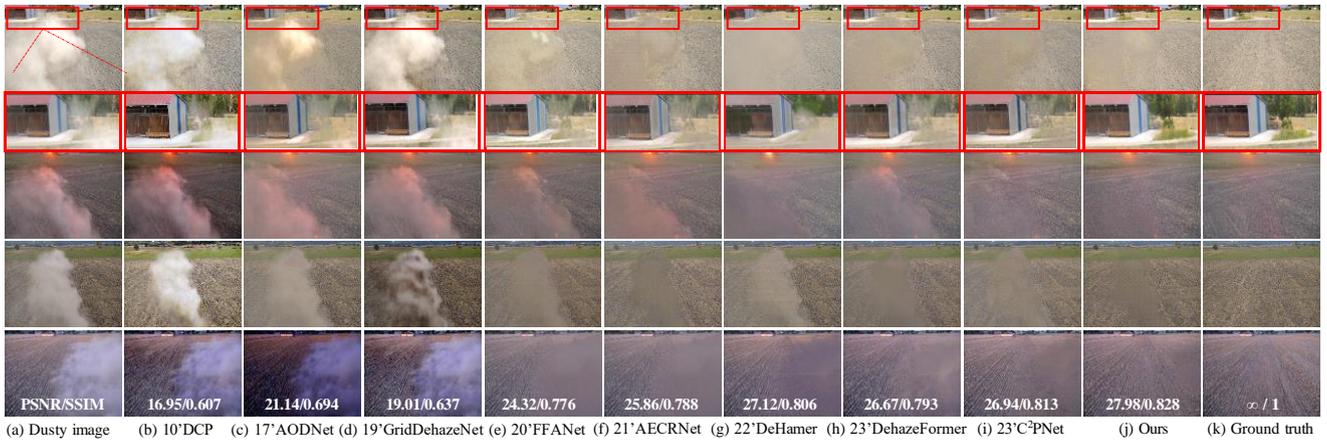

**Figure 8:** Qualitative comparison results of DedustNet and SOTA methods on RB-Dust datasets.

The MS-SSIM loss integrates the variations of resolution and visualization conditions to consider structural differences; compared to other loss functions, the MS-SSIM loss preserves the contrast in the high-frequency region.

Inspired by the current hot research in image dehazing, we adopt perceptual loss (Simonyan and Zisserman (2014)) to promote the perceptual similarity of dimensional spatial features and perceive the image from a high dimension.

## 4. Experimental setup

This section is about our dataset preparations. We introduce the agricultural dust datasets we use in Section 4.1 and our datasets expansion way in Section 4.2. In Section 4.3, we introduce four real-world hazy datasets to validate the network's generalization ability. Finally, the comparison methods and our training details are embodied in Section 4.4.

### 4.1. Dust dataset for training and test

To the best of our knowledge, RB-Dust Dataset (Buckel et al. (2023)) is the only publicly available agricultural landscape dusting dataset, so we conducted experiments on it, which consists of 200 images with 1920 × 1080 pixels. We selected 180 images in the overall dataset, cropped the original pixels to a size of 500×500, and data enhancement operations (including horizontal flipping, rotation, etc.), and used the enhanced dataset for training. However, it should be noted that we utilize the original pixel size of the remaining 20 images for testing.

### 4.2. Real-world datasets expansion

We randomly cropped the original images in the RB-Dust dataset into square patches of 512 × 512 pixels; these patches are not the same for every epoch. To augment the training data, we implemented random rotations (90, 180, or 270 degrees) and random horizontal flips when processing the training data. This step allows this small real-world dataset to be expanded into larger datasets, which can be more suitable for training the data-driven methods.

### 4.3. Haze datasets for generality analysis

We have performed generality analysis on the following four real-world hazy datasets: NTIRE 2018 image dehazing dataset (I-Haze (Ancuti, Ancuti, Timofte and De Vleeschouwer (2018a))), the outdoor NTIRE 2018 image dehazing dataset (O-Haze (Ancuti, Ancuti, Timofte and De Vleeschouwer (2018b))), a benchmark for image dehazing with densehaze and haze-free images (Dense-Haze (Ancuti, Ancuti, Sbert and Timofte (2019))), and the NTIRE 2020 dataset for non-homogeneous dehazing challenge (NH-Haze (Ancuti, Ancuti, Vasluianu and Timofte (2020))).

**I-Haze (Ancuti et al. (2018a)) and O-Haze (Ancuti et al. (2018b)):** They contain 25 and 35 hazy images (size 2833 × 4657 pixels) respectively for training. Both datasets contain five hazy images for validation.

**Dense-Haze (Ancuti et al. (2019)):** It contains 45 hazy images (size 1200 × 1600 pixels) for training five hazy images for validation and five more for testing with their corresponding ground truth images.

**NH-Haze (Ancuti et al. (2020))**

### 4.4. Comparison methods and training details

Because the effectiveness of image dehazing algorithms for the dust removal task has been verified in the work (Buckel et al. (2023)), therefore, we selected the state-of-the-art (SOTA) dehazing methods and our algorithms to be trained on this dataset with the same experimental setting. To make a comprehensive and objective comparison, we mainly select some SOTA methods for comparison according to the three main categories of image dehazing methods, including prior-based methods and hazy-to-clear image translation-based methods.

We conducted a comparative experiment on GeForce RTX 4090 using PyTorch 1.11.0. Adam optimizer is adopted, the initial learning rate is set to 0.0001, betas = (0.9, 0.999),

**Table 1**

Quantitative comparisons among DedustNet and SOTA methods on RB-Dust dataset. Indicators marked with ↑ indicate higher and better data, and ↓ indicate lower and better. **Bold** and underline mark the best and second-best methods, respectively.

| Methods | RB-Dust dataset | | | | |
|---|---|---|---|---|---|
| | PSNR↑ | SSIM↑ | Entropy↑ | NIQE↓ | FADE↓ |
| TAPAMI'10 DCP He et al. (2010) | 16.95 | 0.607 | 6.2502 | 4.9805 | 0.9230 |
| ICCV'17 AODNet Li et al. (2017) | 21.04 | 0.694 | 6.3331 | 4.8651 | 0.8943 |
| ICCV'19 GridDehazeNet Liu et al. (2019) | 19.01 | 0.637 | 6.2824 | 4.5640 | 0.8432 |
| CVPR'20 MSBDN Dong et al. (2020a) | 20.73 | 0.698 | 6.8354 | 4.1913 | 0.7546 |
| AAAI'20 FFANet Qin et al. (2020) | 24.32 | 0.776 | 6.6523 | 3.7932 | 0.6749 |
| CVPR'21 AECRNet Wu et al. (2021) | 25.86 | 0.788 | <u>7.2467</u> | 3.6064 | 0.5980 |
| CVPR'22 DeHamer Guo et al. (2022) | <u>27.12</u> | 0.806 | 7.0986 | 3.6287 | 0.4887 |
| TIP'23 DehazeFormer-B Song et al. (2023) | 26.67 | 0.793 | 7.1526 | 3.4691 | 0.4652 |
| CVPR'23 C$^2$PNet Zheng et al. (2023) | 26.94 | <u>0.813</u> | 7.2211 | <u>3.4577</u> | <u>0.4501</u> |
| **DedustNet (Ours)** | **27.98** | **0.828** | **7.5472** | **3.3450** | **0.4266** |

the batch size is 16, the crop size is 256 × 256, and the total number of the epoch is 150.

## 5. Experimental results and analysis

This section demonstrates the qualitative and quantitative comparison results among DedustNet and SOTA methods in Section 5.1 and generality analysis for DedustNet in Section 5.2. In Section 5.3, we conduct the application test among DedustNet and SOTA methods. In Section 5.4, we show the parameters and runtime analysis. Finally, the ablation study is labeled in Section 5.5.

### 5.1. Qualitative and quantitative comparisons
#### 5.1.1. Qualitative comparisons with SOTA methods

Fig. 8 illustrates that DCP, AODNet, GridDehazeNet, and MSBDN produce incomplete removal results when faced with non-homogeneous dust, especially AODNet and MSBDN deepen the color of the image itself due to over-enhancement; FFANet, AECRNet, and DehazeFormer can remove most of the dust, but there is still some dust left; DeHamer and C$^2$PNet bring good overall results, but there is a certain degree of detail loss. In contrast, our dedusting results are closest to the ground truth. DedustNet maximizes the removal of non-homogeneous dust without sacrificing detailed information and obtains better visualization results.

#### 5.1.2. Qualitative comparisons with SOTA methods

Table 1 demonstrates that compared to the SOTA method, our DedustNet with lower NIQE is ahead in PSNR, SSIM, and Entropy by 0.86dB, 0.015, and 0.3005, relatively. Entropy and NIQE are the two non-referenced indicators; higher Entropy metrics prove that the image contains more helpful information, and lower NIQE metrics indicate better image quality. These results demonstrate that the output image of DedustNet has high quality.

**Table 2**

Quantitative comparisons on computational efficiency among DedustNet and SOTA methods, where the floating-point operations and inference time are measured on RGB image with a resolution of 256 × 256. Data marked with - is not available.

| Methods | Overhead | | |
|---|---|---|---|
| | #Param↓ | #FLOPs | Runtime↓ |
| TAPAMI'10 DCP He et al. (2010) | - | - | - |
| ICCV'17 AODNet Li et al. (2017) | 0.002M | 0.115G | 0.316ms |
| ICCV'19 GridDehazeNet Liu et al. (2019) | 0.956M | 21.49G | 15.35ms |
| CVPR'20 MSBDN Dong et al. (2020a) | 31.35M | 41.54G | 9.826ms |
| AAAI'20 FFANet Qin et al. (2020) | 4.456M | 287.8G | 47.98ms |
| CVPR'21 AECRNet Wu et al. (2021) | 2.611M | 52.20G | - |
| CVPR'22 DeHamer Guo et al. (2022) | 132.45M | 48.93G | 14.12ms |
| TIP'23 DehazeFormer-B Song et al. (2023) | 2.514M | 25.79G | 20.79ms |
| CVPR'23 C$^2$PNet Zheng et al. (2023) | 7.17M | - | - |
| **DedustNet (Ours)** | **1.866M** | **4.08G** | **15.96ms** |

### 5.2. Generality analysis for DedustNet

Fig. 10 and Fig. 11 illustrate that DedustNet also achieves satisfactory results on four real-world haze datasets (I-Haze, O-Haze, Deense-Haze, and NH-Haze datasets) compared to SOTA methods, which illustrates that DedustNet removes more dense and non-homogeneous dust and retains more textural detail, demonstrating the promising robustness and generalization ability of DedustNet.

### 5.3. Application test

The SIFT algorithm (Lowe (2004)) is used to detect and describe the matching of feature points between different images by extracting the local features of the image. This approach has a wide range of applications in the fields of target recognition and target tracking. In this section, we perform a feature point matching test to evaluate the performance of DedustNet. We selected the last four years of SOTA methods for comparison. Fig. 13 shows more of the application test results among our method and SOTA methods, and we can observe that DedustNet has the highest number of matching points. The application test shows that DedustNet exhibits better performance in computer vision-related applications.

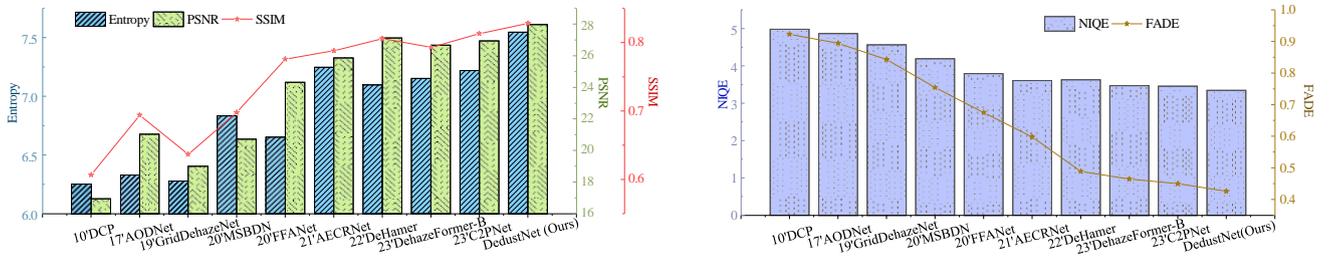

**Figure 9:** Quantitative comparison results between DedustNet and SOTA methods on the RB-Dust dataset. Higher metrics are better in the left image; lower metrics are better in the right image.

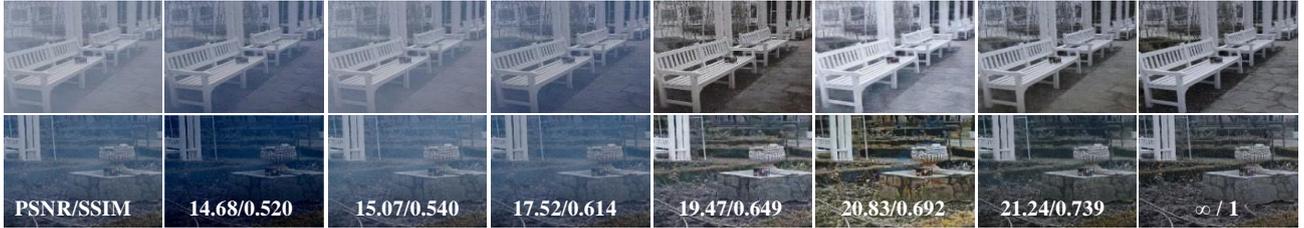

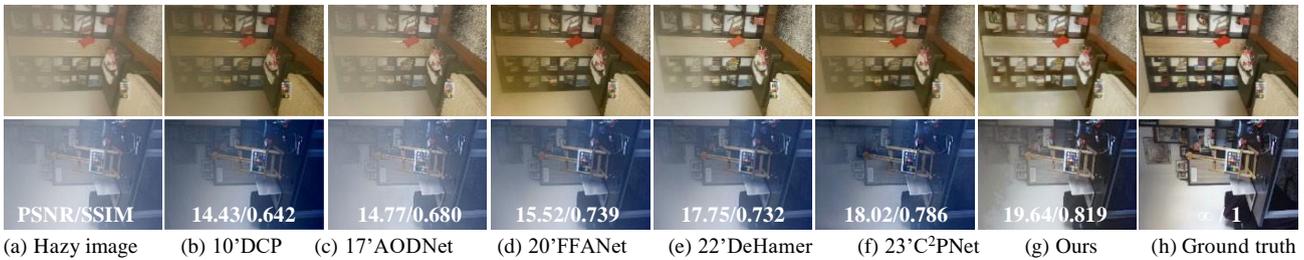

**Figure 10:** Qualitative comparison of DedustNet and SOTA methods on four real-world haze datasets (I-Haze, O-Haze, Dense-Haze, and NH-Haze datasets).

**Table 3**

Quantitative results of ablation study on losses function of DedustNet on the RB-Dust dataset, "w/o" means without.

| | Loss function | | Results on RB-Dust dataset | |
|---|---|---|---|---|
| $\ell_1$ | MS-SSIM loss | perceptual loss | PSNR↑ | SSIM↑ |
| ✓ | | | 23.79 | 0.781 |
| ✓ | ✓ | | 26.49 | 0.815 |
| ✓ | ✓ | ✓ | **27.98** | **0.828** |

### 5.4. Parameters and runtime analysis

Table 2 demonstrates that our method has a significant advantage over the SOTA methods in parameters metric with a slight operational complexity. Still, our approach does not have a substantial advantage in inference time because the wavelet transform process takes a certain amount of time. However, DedustNet outperforms the SOTA methods in quantitative and qualitative comparisons. Therefore, our proposed method has a significant advantage in a comprehensive view of the number of network parameters, model complexity, and overall network performance.

### 5.5. Ablation study

We conducted an ablation study on the RB-Dust dataset. We can intuitively observe in Fig. 14 that each module designed and used contributes to the dust removal performance of DedustNet (where w/o is an abbreviation for without). Precisely, our proposed DWTFormer Block with SFAS can remove most of the dense dust, the CIFM module serves as a link between the encoding and decoding of the network features, smoothly handling most of the shadow residues and recovering richer image details, and the DCM module we adopted to serve as a transition module, bring a very obvious gain to the overall dedusting performance.

The results of the ablation experiments on loss function are placed in Table 3. The $\ell_1$ acts as a widely used loss function, which retains the color and brightness of the raw images. The MS-SSIM loss provided PSNR and SSIM values of 2.7 and 0.034, respectively, for the overall network. The perceptual loss provided PSNR and SSIM values of 1.49 and 0.13, respectively. These results demonstrate the gain that each loss function brings to the dedusting performance of DedustNet.

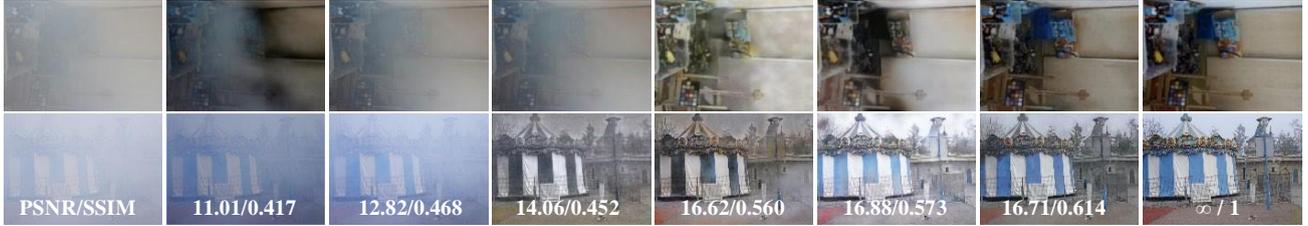
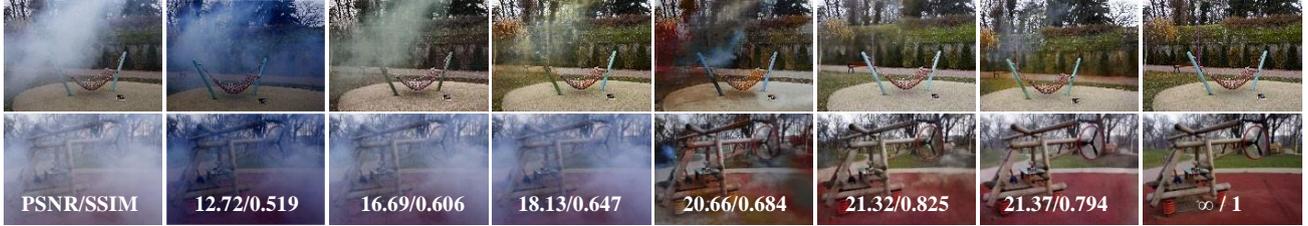

**Figure 11:** Qualitative comparison of DedustNet and SOTA methods on four real-world haze datasets (I-Haze, O-Haze, Dense-Haze, and NH-Haze datasets).

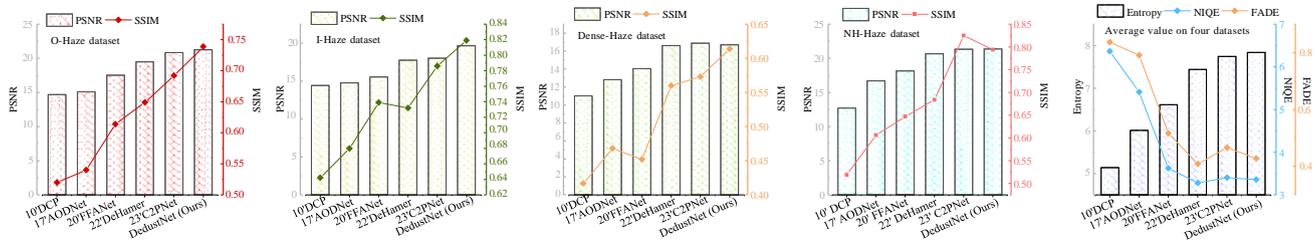

**Figure 12:** Quantitative comparison results between DedustNet and SOTA methods on four real-world fog benchmark datasets.

## 6. Conclusion

This paper focuses on image dedusting in agriculture, which is highly practical and significant but has yet to receive yet to receive widespread attention. A frequency-dominated Swin Transformer-based wavelet network (DedustNet) for real-world agricultural landscape dedusting is proposed as a solution. We introduce frequency-dominated Swin Transformer-based blocks (DWTFormer and IDWTFormer) with SFAS to enhance the performance of Swin Transformer in dealing with complex dusty backgrounds, which effectively recovers details such as the structure and texture of the image. Additionally, our proposed CIFM captures global and long-range feature relationships, while the DCM utilizes wavelet-guided dilated convolutions to extract contextual information at multiple scales. Our proposed DedustNet has shown superior performance and more reliable results in agricultural image dedusting with better generalization capability when compared to SOTA methods. Our research provides novel solutions to image dusting tasks, and we will build more high-quality datasets for the agricultural landscape dusting task in our further work.

## 7. Limitations and discussion

Although DedustNet achieves satisfactory results on image dust removal tasks in agriculture, we still find that the output of DedustNet still needs complete removal or artifacts in Fig. 15 when faced with dense dust or more complex dust backgrounds. This is because there are currently too few dust datasets from natural agriculture available for training, and these inadequate datasets limit the expressive power of our model. This is what we will work on next. We will propose more and better quality datasets of agricultural images in our subsequent work.

## 8. Acknowledgements

This work was supported in part by the Applied Basic Research Project of Department of Science & Technology of Liaoning province under Grant 2022JH2/101300274, in part by the Educational Department of Liaoning Province under Grant No.LJKMZ20220679.

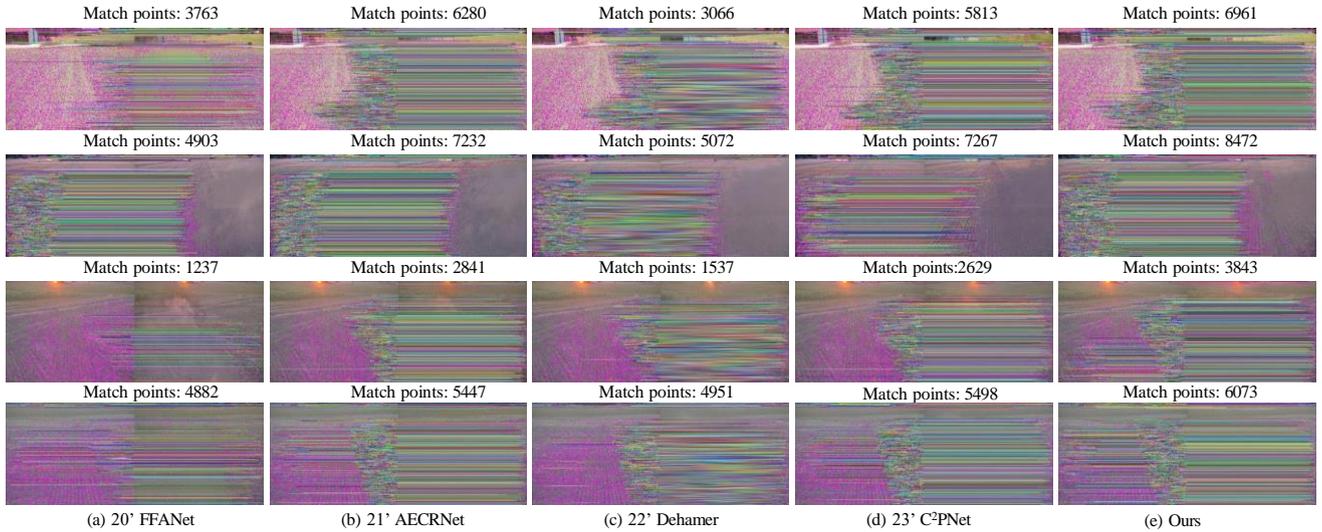

**Figure 13:** Application test results of DedustNet and SOTA methods on RB-Dust dataset. The purple dots represent feature points, and the horizontal lines represent the matching of feature points between the dehazed result by different methods (right one) and a clear reference image (left one); the denser the matching lines are, the higher the degree of feature matching.

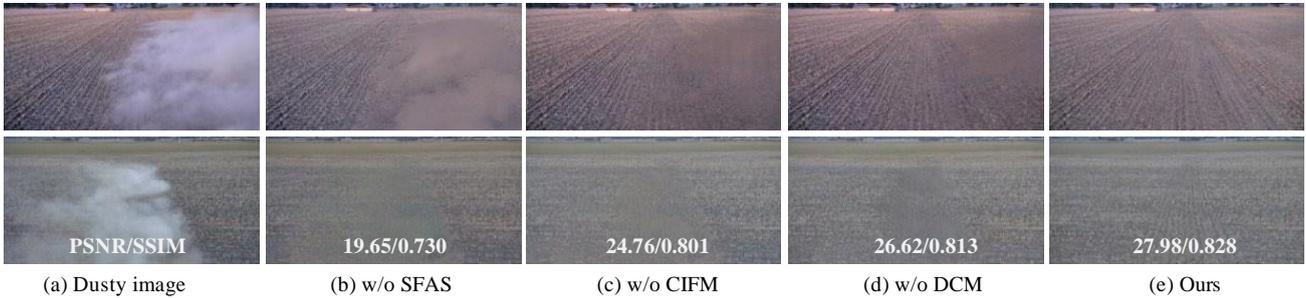

**Figure 14:** Visualization comparative results on RB-Dust dataset for different variants of DedustNet. "w/o" means without, and the data on the images represent the average values on the RB-Dust dataset during the ablation study.

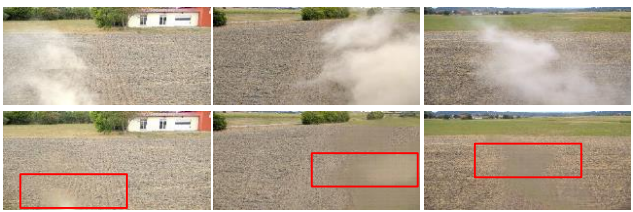

**Figure 15:** Artifacts in DedustNet when processing some images with dense dust. The red frame represents a zoomed-in detail.

## CRediT authorship contribution statement

**Shengli Zhang:** Conceptualization of this study, Methodology, Software, Writing - Original Draft. **Zhiyong Tao:** Supervision, Writing - Review Editing. **Sen Lin:** Supervision, Writing - Review Editing.

## References

Ancuti, C., Ancuti, C.O., Timofte, R., De Vleeschouwer, C., 2018a. I-haze: a dehazing benchmark with real hazy and haze-free indoor images, in: Advanced Concepts for Intelligent Vision Systems: 19th International Conference, ACIVS 2018, Poitiers, France, September 24–27, 2018, Proceedings 19, Springer. pp. 620–631.

Ancuti, C.O., Ancuti, C., Sbert, M., Timofte, R., 2019. Dense-haze: A benchmark for image dehazing with dense-haze and haze-free images, in: 2019 IEEE international conference on image processing (ICIP), IEEE. pp. 1014–1018.

Ancuti, C.O., Ancuti, C., Timofte, R., De Vleeschouwer, C., 2018b. O-haze: a dehazing benchmark with real hazy and haze-free outdoor images, in: Proceedings of the IEEE conference on computer vision and pattern recognition workshops, pp. 754–762.

Ancuti, C.O., Ancuti, C., Vasluianu, F.A., Timofte, R., 2020. Ntire 2020 challenge on nonhomogeneous dehazing, in: Proceedings of the IEEE/CVF Conference on Computer Vision and Pattern Recognition Workshops, pp. 490–491.

Buckel, P., Oksanen, T., Dietmueller, T., 2023. Rb-dust-a reference-based dataset for vision-based dust removal, in: Proceedings of the IEEE/CVF Conference on Computer Vision and Pattern Recognition, pp. 1140–1149.

Cai, B., Xu, X., Jia, K., Qing, C., Tao, D., 2016. Dehazenet: An end-to-end system for single image haze removal. IEEE Transactions on Image Processing 25, 5187–5198.

Chen, L.C., Zhu, Y., Papandreou, G., Schroff, F., Adam, H., 2018. Encoder-decoder with atrous separable convolution for semantic image segmentation, in: Proceedings of the European conference on computer vision (ECCV), pp. 801–818.

Claypoole, R., Baraniuk, R., Nowak, R., 1998. Adaptive wavelet transforms via lifting, in: Proceedings of the 1998 IEEE International Conference on Acoustics, Speech and Signal Processing, ICASSP '98 (Cat. No.98CH36181), pp. 1513–1516 vol.3. doi:10.1109/ICASSP.1998.681737.

Das, S.D., Dutta, S., 2020. Fast deep multi-patch hierarchical network for nonhomogeneous image dehazing, in: Proceedings of the IEEE/CVF conference on computer vision and pattern recognition workshops, pp. 482–483.

Dong, H., Pan, J., Xiang, L., Hu, Z., Zhang, X., Wang, F., Yang, M.H., 2020a. Multi-scale boosted dehazing network with dense feature fusion, in: Proceedings of the IEEE/CVF conference on computer vision and pattern recognition, pp. 2157–2167.

Dong, Y., Liu, Y., Zhang, H., Chen, S., Qiao, Y., 2020b. Fd-gan: Generative adversarial networks with fusion-discriminator for single image dehazing, in: Proceedings of the AAAI Conference on Artificial Intelligence, pp. 10729–10736.

Dosovitskiy, A., Beyer, L., Kolesnikov, A., Weissenborn, D., Zhai, X., Unterthiner, T., Dehghani, M., Minderer, M., Heigold, G., Gelly, S., et al., 2020. An image is worth 16x16 words: Transformers for image recognition at scale. arXiv preprint arXiv:2010.11929 .

Fattal, R., 2014. Dehazing using color-lines. ACM transactions on graphics (TOG) 34, 1–14.

Fu, M., Liu, H., Yu, Y., Chen, J., Wang, K., 2021. Dw-gan: A discrete wavelet transform gan for nonhomogeneous dehazing, in: Proceedings of the IEEE/CVF Conference on Computer Vision and Pattern Recognition, pp. 203–212.

Guo, C.L., Yan, Q., Anwar, S., Cong, R., Ren, W., Li, C., 2022. Image dehazing transformer with transmission-aware 3d position embedding, in: Proceedings of the IEEE/CVF Conference on Computer Vision and Pattern Recognition, pp. 5812–5820.

Guo, T., Seyed Mousavi, H., Huu Vu, T., Monga, V., 2017. Deep wavelet prediction for image super-resolution, in: Proceedings of the IEEE conference on computer vision and pattern recognition workshops, pp. 104–113.

He, K., Sun, J., Tang, X., 2010. Single image haze removal using dark channel prior. IEEE transactions on pattern analysis and machine intelligence 33, 2341–2353.

Jiang, K., Wang, Z., Yi, P., Jiang, J., Xiao, J., Yao, Y., 2018. Deep distillation recursive network for remote sensing imagery super-resolution. Remote Sensing 10, 1700.

Jiang, K., Wang, Z., Yi, P., Wang, G., Lu, T., Jiang, J., 2019. Edge-enhanced gan for remote sensing image superresolution. IEEE Transactions on Geoscience and Remote Sensing 57, 5799–5812.

Li, B., Peng, X., Wang, Z., Xu, J., Feng, D., 2017. Aod-net: All-in-one dehazing network, in: Proceedings of the IEEE international conference on computer vision, pp. 4770–4778.

Li, H., Li, J., Zhao, D., Xu, L., 2021. Dehazeflow: Multi-scale conditional flow network for single image dehazing, in: Proceedings of the 29th ACM International Conference on Multimedia, pp. 2577–2585.

Li, Z., Zheng, C., Shu, H., Wu, S., 2022. Dual-scale single image dehazing via neural augmentation. IEEE Transactions on Image Processing 31, 6213–6223.

Liu, P., Zhang, H., Zhang, K., Lin, L., Zuo, W., 2018. Multi-level wavelet-cnn for image restoration, in: Proceedings of the IEEE conference on computer vision and pattern recognition workshops, pp. 773–782.

Liu, W., Yan, Q., Zhao, Y., 2020. Densely self-guided wavelet network for image denoising, in: Proceedings of the IEEE/CVF Conference on Computer Vision and Pattern Recognition Workshops, pp. 432–433.

Liu, X., Ma, Y., Shi, Z., Chen, J., 2019. Griddehazenet: Attention-based multi-scale network for image dehazing, in: Proceedings of the IEEE/CVF international conference on computer vision, pp. 7314–7323.

Liu, Y., Zhu, L., Pei, S., Fu, H., Qin, J., Zhang, Q., Wan, L., Feng, W., 2021a. From synthetic to real: Image dehazing collaborating with unlabeled real data, in: Proceedings of the 29th ACM international conference on multimedia, pp. 50–58.

Liu, Z., Lin, Y., Cao, Y., Hu, H., Wei, Y., Zhang, Z., Lin, S., Guo, B., 2021b. Swin transformer: Hierarchical vision transformer using shifted windows, in: Proceedings of the IEEE/CVF international conference on computer vision, pp. 10012–10022.

Lowe, D.G., 2004. Distinctive image features from scale-invariant keypoints. International journal of computer vision 60, 91–110.

McCartney, E.J., 1976. Optics of the atmosphere: scattering by molecules and particles. New York .

Mehta, A., Sinha, H., Mandal, M., Narang, P., 2021. Domain-aware unsupervised hyperspectral reconstruction for aerial image dehazing, in: Proceedings of the IEEE/CVF winter conference on applications of computer vision, pp. 413–422.

Narasimhan, S.G., Nayar, S.K., 2002. Vision and the atmosphere. International journal of computer vision 48, 233.

Nayar, S.K., Narasimhan, S.G., 1999. Vision in bad weather, in: Proceedings of the seventh IEEE international conference on computer vision, IEEE. pp. 820–827.

Petit, O., Thome, N., Rambour, C., Themyr, L., Collins, T., Soler, L., 2021. U-net transformer: Self and cross attention for medical image segmentation, in: Machine Learning in Medical Imaging: 12th International Workshop, MLMI 2021, Held in Conjunction with MICCAI 2021, Strasbourg, France, September 27, 2021, Proceedings 12, Springer. pp. 267–276.

Qin, X., Wang, Z., Bai, Y., Xie, X., Jia, H., 2020. Ffa-net: Feature fusion attention network for single image dehazing, in: Proceedings of the AAAI conference on artificial intelligence, pp. 11908–11915.

Rasti, P., Uiboupin, T., Escalera, S., Anbarjafari, G., 2016. Convolutional neural network super resolution for face recognition in surveillance monitoring, in: Articulated Motion and Deformable Objects: 9th International Conference, AMDO 2016, Palma de Mallorca, Spain, July 13-15, 2016, Proceedings 9, Springer. pp. 175–184.

Ren, W., Liu, S., Zhang, H., Pan, J., Cao, X., Yang, M.H., 2016a. Learning to see through fog, in: Proceedings of the IEEE Conference on Computer Vision and Pattern Recognition, pp. 1–9.

Ren, W., Liu, S., Zhang, H., Pan, J., Cao, X., Yang, M.H., 2016b. Single image dehazing via multi-scale convolutional neural networks, in: Computer Vision–ECCV 2016: 14th European Conference, Amsterdam, The Netherlands, October 11-14, 2016, Proceedings, Part II 14, Springer. pp. 154–169.

Ren, W., Ma, L., Zhang, J., Pan, J., Cao, X., Liu, W., Yang, M.H., 2018. Gated fusion network for single image dehazing, in: Proceedings of the IEEE conference on computer vision and pattern recognition, pp. 3253–3261.

Ronneberger, O., Fischer, P., Brox, T., 2015. U-net: Convolutional networks for biomedical image segmentation, in: Medical Image Computing and Computer-Assisted Intervention–MICCAI 2015: 18th International Conference, Munich, Germany, October 5-9, 2015, Proceedings, Part III 18, Springer. pp. 234–241.

Shen, H., Zhang, C., Li, H., Yuan, Q., Zhang, L., 2020. A spatial–spectral adaptive haze removal method for visible remote sensing images. IEEE Transactions on Geoscience and Remote Sensing 58, 6168–6180.

Simonyan, K., Zisserman, A., 2014. Very deep convolutional networks for large-scale image recognition. arXiv preprint arXiv:1409.1556 .

Singh, A., Bhave, A., Prasad, D.K., 2020. Single image dehazing for a variety of haze scenarios using back projected pyramid network, in: Computer Vision–ECCV 2020 Workshops: Glasgow, UK, August 23–28, 2020, Proceedings, Part IV 16, Springer. pp. 166–181.

Song, Y., He, Z., Qian, H., Du, X., 2023. Vision transformers for single image dehazing. IEEE Transactions on Image Processing .

Vaswani, A., Shazeer, N., Parmar, N., Uszkoreit, J., Jones, L., Gomez, A.N., Kaiser, Ł., Polosukhin, I., 2017. Attention is all you need. Advances in neural information processing systems 30.

Wang, C., Huang, Y., Zou, Y., Xu, Y., 2021a. Fwb-net: front white balance network for color shift correction in single image dehazing via atmospheric light estimation, in: ICASSP 2021-2021 IEEE International Conference on Acoustics, Speech and Signal Processing (ICASSP), IEEE. pp. 2040–2044.

Wang, P., Zhu, H., Huang, H., Zhang, H., Wang, N., 2022. Tms-gan: A twofold multi-scale generative adversarial network for single image dehazing. IEEE Transactions on Circuits and Systems for Video Technology 32, 2760–2772. doi:10.1109/TCSVT.2021.3097713.

Wang, Y., Yu, X., An, D., Wei, Y., 2021b. Underwater image enhancement and marine snow removal for fishery based on integrated dual-channel neural network. Computers and Electronics in Agriculture 186, 106182.

Wang, Z., Yi, P., Jiang, K., Jiang, J., Han, Z., Lu, T., Ma, J., 2018. Multi-memory convolutional neural network for video super-resolution. IEEE Transactions on Image Processing 28, 2530–2544.

Wu, H., Qu, Y., Lin, S., Zhou, J., Qiao, R., Zhang, Z., Xie, Y., Ma, L., 2021. Contrastive learning for compact single image dehazing, in: Proceedings of the IEEE/CVF Conference on Computer Vision and Pattern Recognition, pp. 10551–10560.

Yang, H.H., Fu, Y., 2019. Wavelet u-net and the chromatic adaptation transform for single image dehazing, in: 2019 IEEE International Conference on Image Processing (ICIP), IEEE. pp. 2736–2740.

Yang, H.H., Yang, C.H.H., Tsai, Y.C.J., 2020. Y-net: Multi-scale feature aggregation network with wavelet structure similarity loss function for single image dehazing, in: ICASSP 2020-2020 IEEE International Conference on Acoustics, Speech and Signal Processing (ICASSP), IEEE. pp. 2628–2632.

Zhang, J., Wang, X., Yang, C., Zhang, J., He, D., Song, H., 2018. Image dehazing based on dark channel prior and brightness enhancement for agricultural remote sensing images from consumer-grade cameras. Computers and Electronics in Agriculture 151, 196–206.

Zhao, H., Gallo, O., Frosio, I., Kautz, J., 2016. Loss functions for image restoration with neural networks. IEEE Transactions on computational imaging 3, 47–57.

Zheng, Y., Zhan, J., He, S., Dong, J., Du, Y., 2023. Curricular contrastive regularization for physics-aware single image dehazing, in: Proceedings of the IEEE/CVF Conference on Computer Vision and Pattern Recognition, pp. 5785–5794.

Zhou, C., Teng, M., Han, Y., Xu, C., Shi, B., 2021. Learning to dehaze with polarization. Advances in Neural Information Processing Systems 34, 11487–11500.

Zhu, Q., Mai, J., Shao, L., 2015. A fast single image haze removal algorithm using color attenuation prior. IEEE transactions on image processing 24, 3522–3533.

Zou, W., Jiang, M., Zhang, Y., Chen, L., Lu, Z., Wu, Y., 2021. Sdwnet: A straight dilated network with wavelet transformation for image deblurring, in: Proceedings of the IEEE/CVF international conference on computer vision, pp. 1895–1904.